# Structure and Implementation of New Practical English Textbooks Driven by Artificial Intelligence

Ya Wang [a], Lei Zhang [a,1], Xueguang Yang [a], Bo Chen [a]
[a]*Air Force Logistics Academy, Xuzhou, 221000,China*

**Abstract.** Artificial intelligence is changing the form of applied English materials from fixed paper sequences to adaptive learning systems that can diagnose learners, recommend tasks, and provide formative feedback. This paper studies the structure and application of a new practical English textbook driven by artificial intelligence. A five-layer architecture is proposed: knowledge mapping, learner profiling, task generation, feedback orchestration, and teacher-side governance. A prototype was tested on 186 non-English-major undergraduates for eight weeks of teaching. Compared with a static digital textbook, the proposed system increased the unit completion accuracy from 72.4% to 84.9%, raised the average score for speaking tasks by 10.8 points, and reduced the teacher's correction time by 31.6%. Therefore, an AI-driven textbook can maintain the stability of the curriculum while providing personalised learning paths, rich practice materials and traceable classroom data.

## 1. Introduction

For a long time, the practical English textbook has served as a basic support for curriculum construction and teaching activities; however, its internal design has often been too rigid and chapter-oriented to meet the diverse needs of learners flexibly. Digital publication has improved accessibility, but many electronic textbooks still present the same fixed sequence of vocabulary, dialogue, reading and exercises as printed materials [1]. The defect in actual English teaching is more noticeable because students need multiple instances of contextual practice, prompt correction, and task transfer in their work, study and life abroad [2]. The recent development of artificial intelligence in education provides technical support for textbooks that can assess students' abilities, suggest learning tasks, and offer personalised learning assistance under the guidance of teachers [3].

Existing AI applications for English learning are generally presented as standalone tools, such as automated writing assessment, pronunciation grading, chatbot interaction, vocabulary exercises, or learning data analysis dashboards. Although the above tools are convenient, they cannot address the textbook-level problem of how to combine content and pedagogy, assessment and teacher orchestration into an effective material system [4]. A learner can receive fluent chatbot practice without a stable syllabus, or obtain automated feedback that is not related to the unit goals [5]. Therefore, the main research problem is not whether AI can create exercises, but how an AI-driven practical English textbook should be organised and applied as a reliable engineering artifact.

[1] Corresponding Author: Lei Zhang

This paper studies the structure and implementation of new practical English textbooks driven by artificial intelligence. Establish a multi-level structure that connects curriculum knowledge points, learner-state modelling, adaptive task sequencing, formative feedback and teacher governance. It adds AI functions to the textbooks rather than adding them as an external addition. The eight-week classroom prototype experiment in this study also employed task accuracy, speaking performance, feedback latency and teacher workload as observable indicators. The goal is to offer a feasible design path for institutions requiring intelligent English teaching materials that still need to be transparent, stable and manageable in the classroom.

## 2. Related Work

### *2.1. AI-supported English learning and feedback*

Research on AI-assisted English learning has expanded from isolated grammar checking to a system of dialogue agents, adaptive recommendation, pronunciation diagnosis and automated writing assessment. Research on chatbot-based English learning shows that conversational agents can increase the frequency of interaction and reduce speaking anxiety under carefully designed task constraints [6]. Automated writing evaluation can also offer revision suggestions, lexical feedback and scores for second-language writing [7]. Recently, research on generative artificial intelligence has focused on its applications in flexible examples, role-playing scripts and personalised explanations, while also pointing out that teachers need to verify the accuracy and suitability of these [8].

Although these studies have expanded the range of application and accelerated feedback, most of them focus on tools or activities rather than using textbooks as the primary medium for organised learning. Isolated intelligence can lead to fragmentation in textbook design; that is, learners are given many prompts but do not know what goals these aim at, and teachers find it difficult to connect the system's output with lesson plans. A good practical English textbook should have a stable knowledge map, reusable task templates, observable progress indicators, and consistent editorial policies. AI functions will have educational value only when they are included in these textbook structures.

### *2.2. Gaps in practical English textbook implementation*

The three kinds of implementation gaps are as follows. First, the content granularity is generally too coarse for adaptation. A unit may contain a dialogue, a vocabulary list and exercises, but the system cannot determine which micro-skill has been lost by the learner. Second, the assessment evidence is not linked to the teaching. Scores are collected after practice, but they do not automatically change the next learning path. Third, teacher management is weak. The teacher will review the suggested tasks, note down any students who require individual attention, and modify the AI-generated activities according to their actual situation in the classroom.

Therefore, the new AI-powered practical English textbook should be designed as an organised environment rather than a collection of content. It needs to keep the curriculum logic of a textbook and add diagnostic, adaptive and analytical functions. Based on the above demands, the engineering question is: what architecture can achieve machine-readable textbook content that is pedagogically sound and operationally feasible. The following section presents a five-layer implementation model and its data flow to address the problem above.

## 3. Structure and Implementation of the AI-driven Practical English Textbook

### *3.1. Textbook knowledge map and learner profile*

At the beginning of each unit in the new textbook, a knowledge map is presented that shows communicative functions, vocabulary, grammar, pragmatics and target outcomes. For example, a workplace unit has specific goals, such as confirming a schedule or using polite modal verbs, and all tasks are linked to clear instructional objectives.

Each learner has a profile that records their mastery, error patterns, response time, preferred learning modes and feedback history. Learner readiness is then calculated as:

$$R_i = \alpha C_i + \beta A_i + \gamma E_i + \delta F_i, \alpha + \beta + \gamma + \delta = 1 \tag{1}$$

where $C_i$ denotes concept mastery, $A_i$ denotes task accuracy, $E_i$ denotes engagement stability, and $F_i$ denotes feedback uptake. The coefficients $\alpha, \beta, \gamma$, and $\delta$ are configured by teachers at the course level, preventing the system from imposing hidden pedagogical priorities. Figure 1 shows the structure of the connections among curriculum nodes, learner profiles, AI services, teacher feedback and adaptive textbook pages.

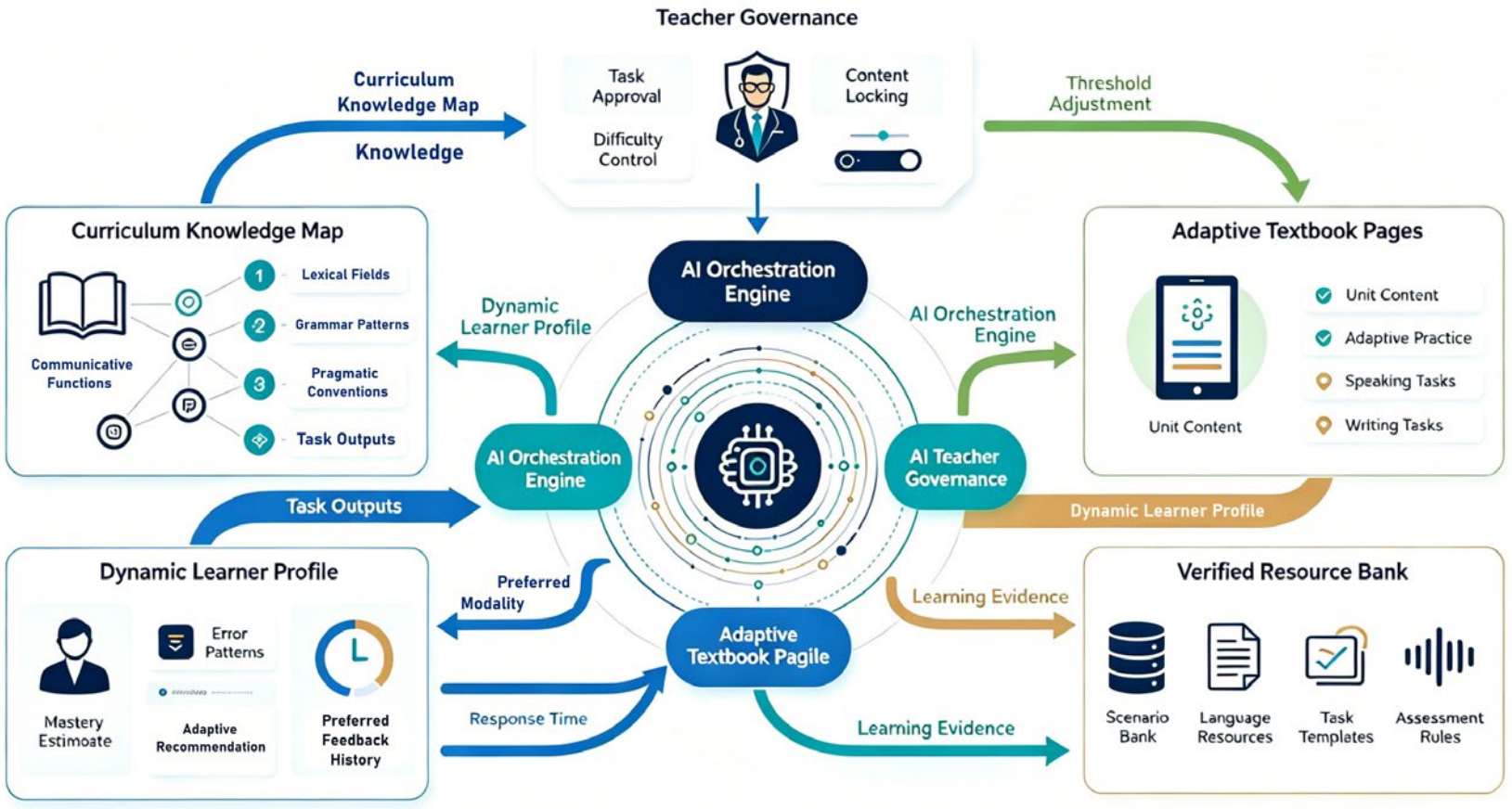

**Figure 1.** Structural framework of the AI-driven practical English textbook

### *3.2. Adaptive Task Generation and Feedback Loop*

The second implementation layer generates adaptive learning problems from the textbook objectives. All task templates have input materials, an expected output, a difficulty tag, a feedback rule and an evidence tag. Instead of generating free-form open-ended activities, a pre-approved template is selected, and then only the scenario variables are filled in from a controlled resource bank. Thus, factual drift is avoided and the generated activities are in line with the syllabus and the purpose of communication.

After each learning session, based on the difference between the expected and actual performance of the learner, update the model parameters.

$$\boldsymbol{\theta}_{t+1} = \boldsymbol{\theta}_t - \eta \nabla_{\boldsymbol{\theta}} \mathcal{L}(\boldsymbol{\theta}_t) \tag{2}$$

where $\boldsymbol{\theta}_t$ represents the adaptive model parameters at iteration $t, \eta$ is the learning rate, and $\mathcal{L}$ is the performance loss function. The update influences subsequent task difficulty, skill coverage, and remediation priority without altering teacher-locked curriculum requirements.

Task updates modify difficulty and focus areas, but respect teacher-locked curriculum settings. The three links in the feedback loop are: real-time error reporting from the system (grammar, vocabulary, pronunciation, format), delayed analysis of recurring errors, and teacher correction for pragmatics and discourse. Figure 2 shows the workflow of diagnostic input and task recommendation to learner response, feedback routing, teacher review and profile update.

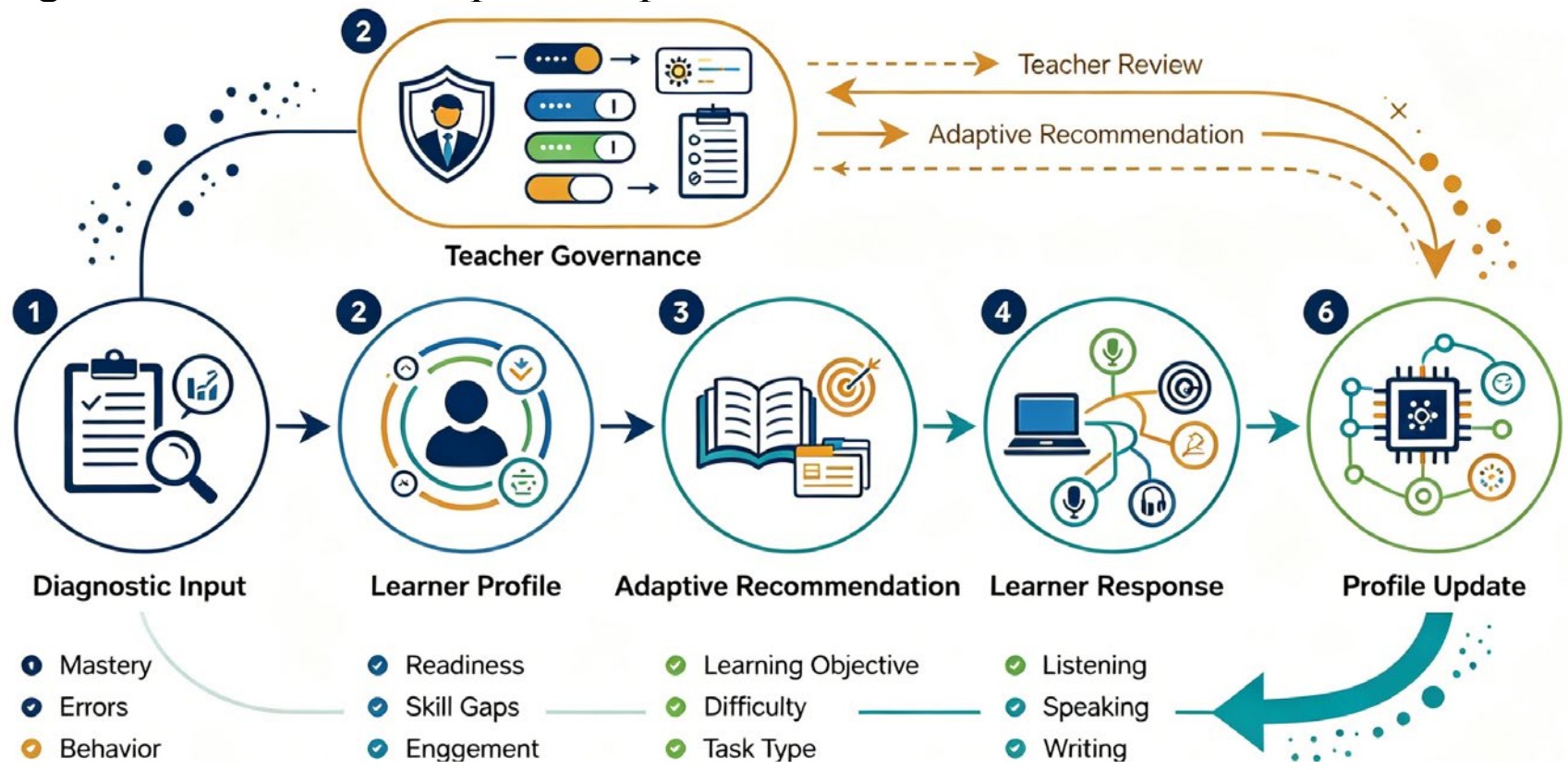

**Figure 2.** Implementation workflow of adaptive task recommendation and feedback routing

### *3.3. Teacher Governance and Textbook Delivery*

Teacher management is the lowest level of implementation. Teachers can lock tasks, approve or reject exercises, set difficulty limits, and export class data according to the progress of the class and curriculum requirements. The system offers instructions and does not replace teachers.

The prototype is a web-based textbook and teacher dashboard. Learners can access adaptive exercises, feedback and previous modifications; teachers will monitor the learning progress of students, organise homework, review generated works, and track feedback frequency. The whole system's efficiency index is as follows:

$$S = 0.30M + 0.25P + 0.20U + 0.15T + 0.10G \tag{3}$$

where $M$ is mastery gain, $P$ is productive-language performance, $U$ is unit completion, $T$ is teacher workload reduction, and $G$ is governance compliance. The weight method does not attribute the fault of poor textbook effectiveness to either test inaccuracy or reduced work. Instructional Quality and learner participation are also included in the evaluation.

## 4. Experimental Data and Analysis

### *4.1. Experimental setting and overall learning performance*

The prototype was used in the four intact practical English classes at a provincial university for testing. A total of 186 first-year non-English-major students participated, and among them, 94 were in the AI-driven textbook group and 92 were in the static digital textbook group. Both groups studied the same eight units of campus service, workplace scheduling, product description, travel assistance, email writing, meeting communication, customer inquiry and presentation opening. The two groups were taught by the same teaching team in the experiment to reduce instructor variation [9].

The pre-test scores of the two groups were statistically the same; the average scores were 66.8/100 for the AI-driven group and 66.1/100 for the control group. After eight weeks, the AI-led group had reached 82.6, and the control group was at 74.3 [10]. Figure 3 shows the results of four-unit clusters, and it can be seen that the AI-driven group achieved an 84.9% completion accuracy, 81.7% listening-response accuracy, 79.4% scenario writing accuracy, and 76.8% speaking-task accuracy; the corresponding control values were 72.4%, 70.2%, 68.1% and 66.0%, respectively. Scenario writing had a relatively large gain, and adaptive feedback repeatedly pointed out missing contextual information [11]; Task-level logs showed that this deficiency was not uniform among all units. The first two units of both groups mainly contained recognition-oriented exercises, and therefore, the difference was only 4.7 percentage points. From Unit 4 onwards, the AI-driven group received more micro-tasks on request forms, confirmation phrases, and short-spoken responses; the control group proceeded through the same content sequence without remediation. By Unit 8, the AI-led group had produced 1.84 valid work products per learner per week and the control group had produced 1.21. The relatively high output density of the textbooks reduced their function as reading materials and increased their role in organised practice.

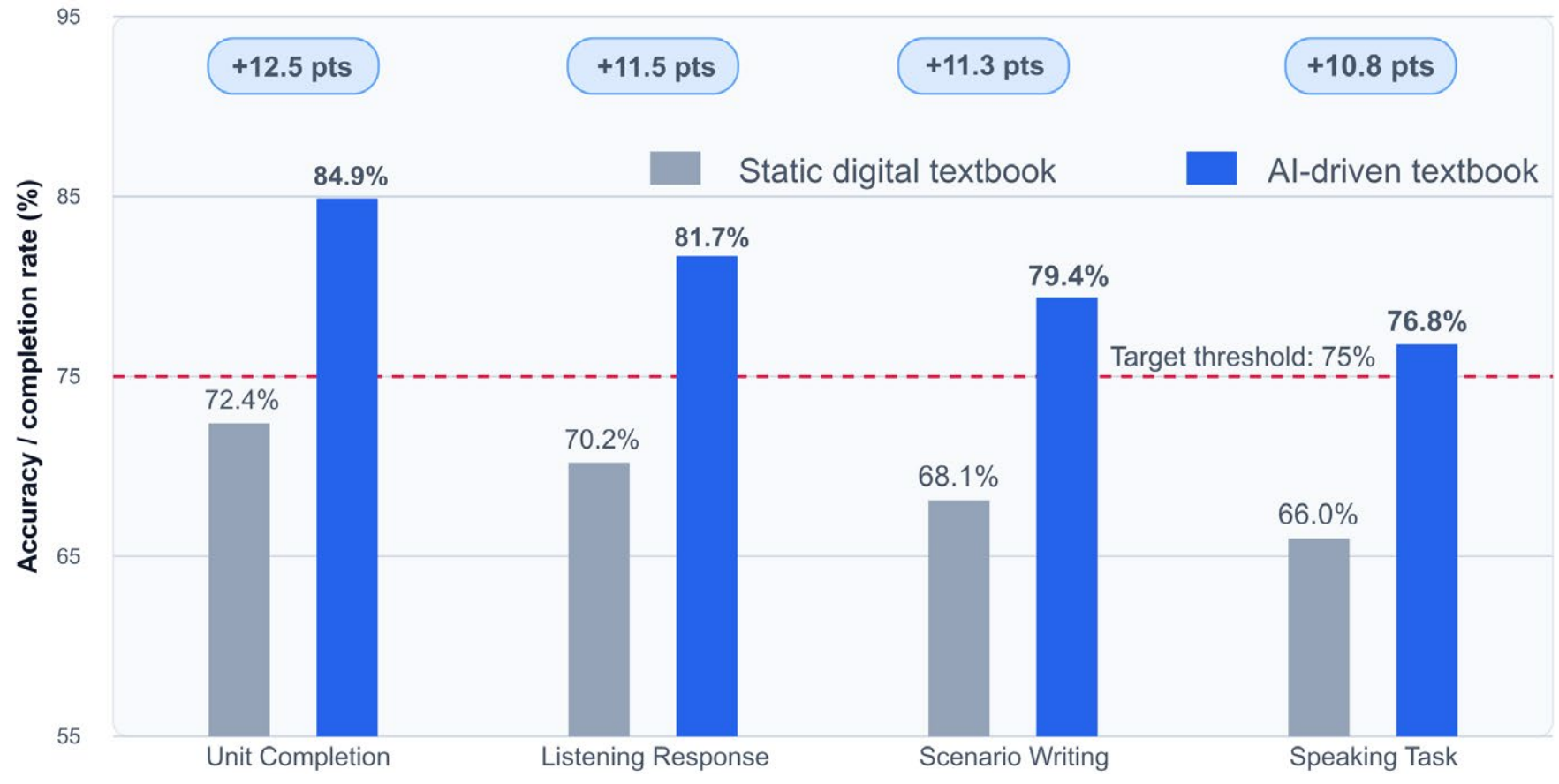

**Figure 3.** Comparative learning performance across four practical English task clusters

*4.2. Feedback efficiency and learner progression*

The first few indices of feedback are response time, approval rate of modifications, number of subsequent errors, and teachers' correction speed. The AI textbook cut the median feedback latency for basic errors from 26.5 hours to 3.2 minutes promptly [12]. Rapid feedback increased the revision acceptance rate from 48.6% (control) to 71.3% (AI) and reduced repeated pragmatic errors per 100 tasks from 18.9 to 10.7 [13].

As shown in Figure 4, after eight weeks, the average mastery index of the AI-guided group increased from 0.43 to 0.82, and that of the control group rose from 0.42 to 0.68. The gap increased after Week 4 due to the personalised remedial tasks available for learners' profiles [14]. Teacher correction time dropped from 7.9 hours in Week 1 to 4.2 hours by Week 8, and this was a 31.6% decrease compared with the control group [15].

Analysis of the feedback problems showed: lexical misuse (34.5%), grammar (27.8%), pragmatic mismatch (21.6%) and task-format (16.1%). The first two were promptly corrected in the system; the pragmatic errors required more teacher intervention, such as revising the complaint response and organising a meeting. Based on the above analysis, the three channels of feedback will be used: automatic correction of local errors and teacher review for discourse and appropriateness.

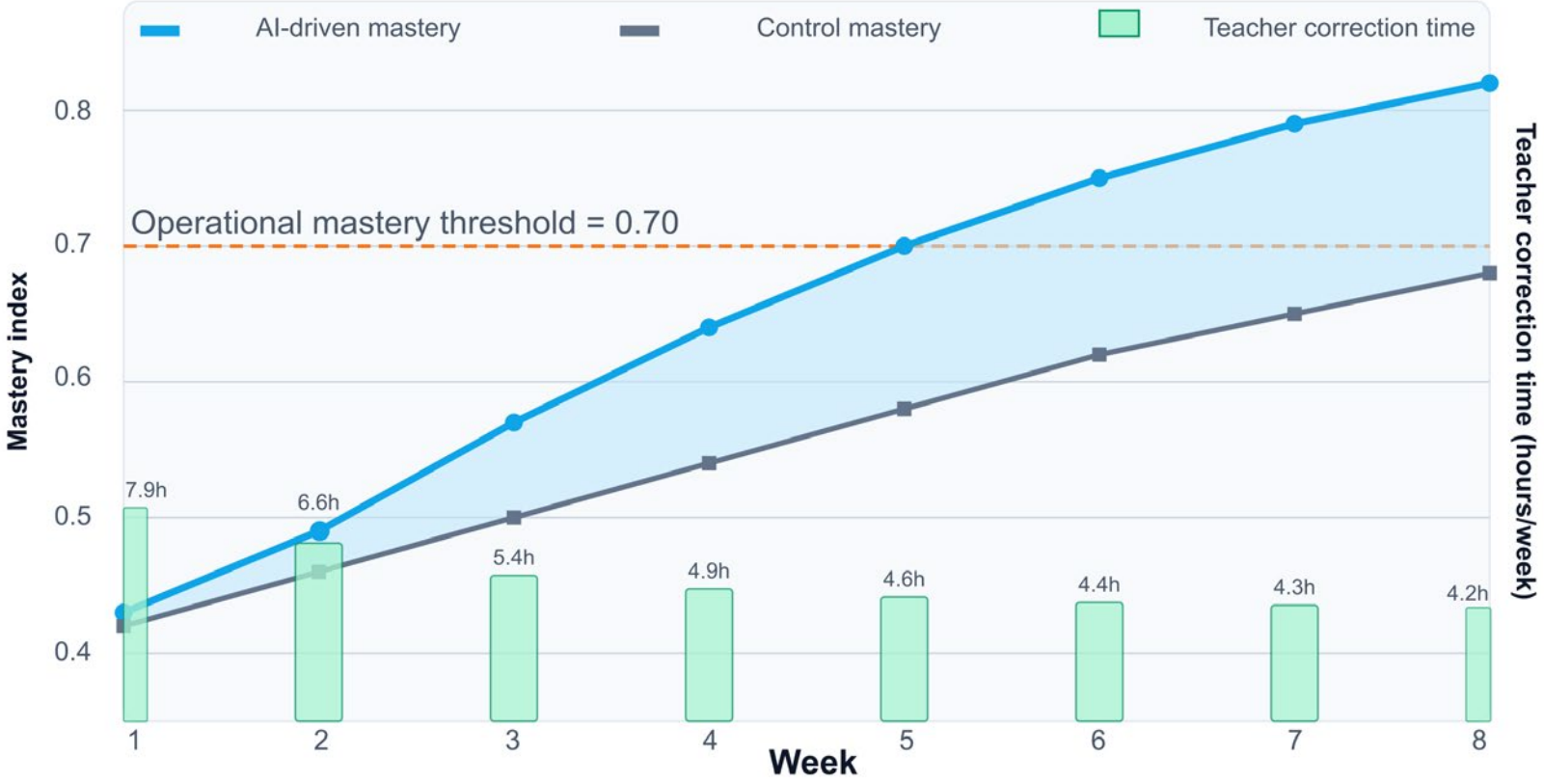

**Figure 4.** Weekly mastery index and teacher correction time during the eight-week experiment

*4.3. Robustness, usability, and implementation risk analysis*

Group students by their pre-test readiness to evaluate convenience for all learners. The improvement after the post-test was most pronounced in the low-readiness group, with an increase of 18.4 points; the medium and high groups increased by 15.2 and 10.1 points, respectively [16]. Thus, adaptive sequencing is suitable for those who need more assistance.

Assess usability using a 5-point scale and logs. Learners rated the following the highest: task clarity (4.31), feedback usefulness (4.26), and interface consistency (4.18). Teacher-side governance scored 4.07 and was lower than learner-side indicators because it lacked more batch-editing functions [17]. As shown in Figure 5, the implementation indicators were: learner satisfaction 86.2%, teacher acceptance 81.5%, content alignment 88.4%, feedback trust 79.8%, governance compliance 92.1%, and technical stability 95.0%. Feedback trust was the lowest, and some still preferred teacher confirmation for high-level discourse [18].

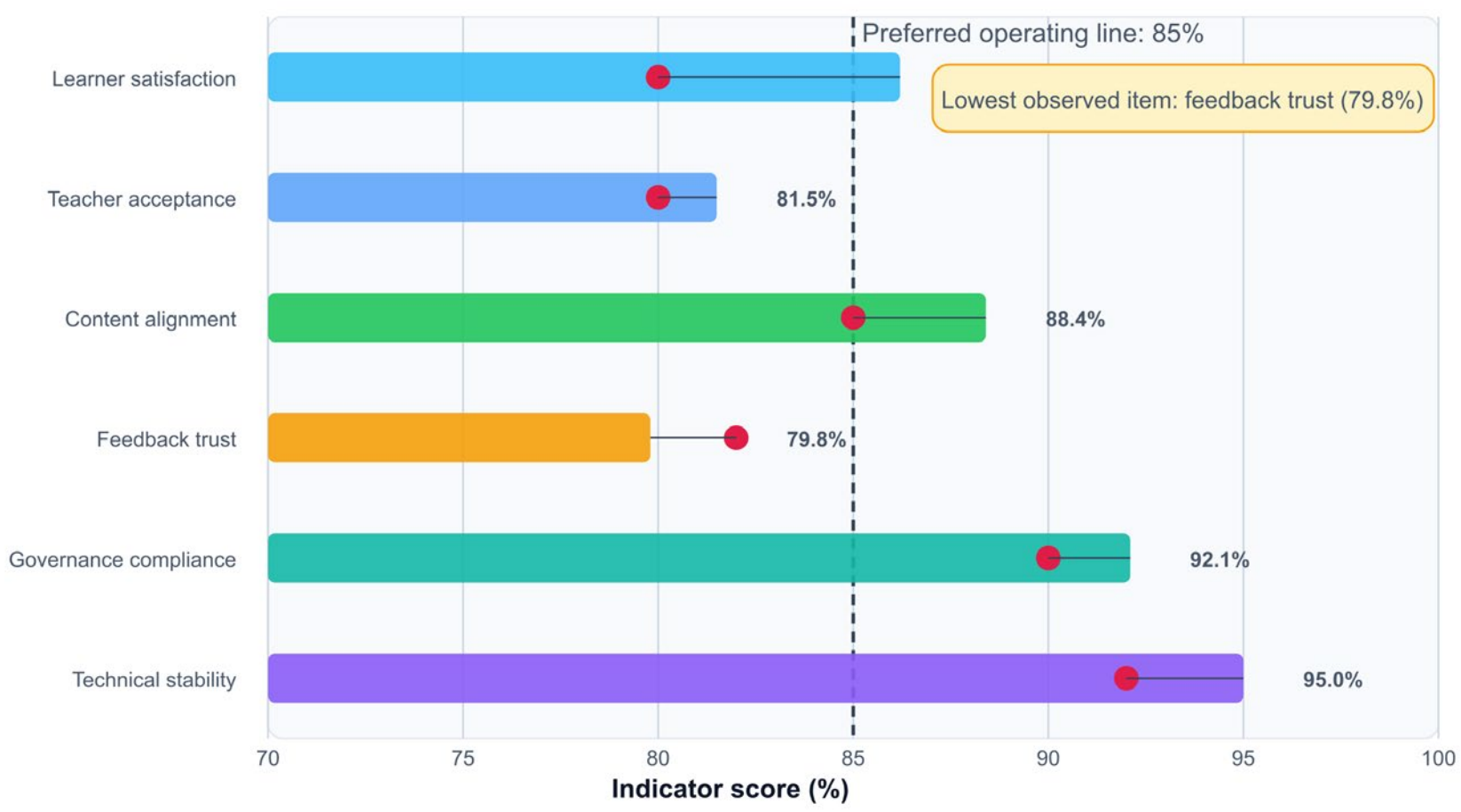

**Figure 5.** Implementation usability and risk-control indicators of the AI-driven textbook

The three problems identified in the risk analysis were: generated tasks should use a validated scenario bank to maintain pedagogical relevance; learner data collection should be limited, and speaking records promptly deleted; teachers need transparent override rights to ensure textbook alignment with course goals, not merely model output [19]. 6.4% of the tasks had to be manually corrected before release due to being too simple or out of context. No high-risk content was released after the revision, and the governance rules have decreased AI uncertainty. Scenario resources, teacher dashboards and thresholds are generally more influential on the implementation quality than the model itself.

## 5. Conclusion

This paper examines the Structure and Application of New Practical English Textbooks Based on Artificial Intelligence. A five-layer architecture was proposed to link knowledge mapping, learner profiling, adaptive task generation, feedback orchestration and teacher governance. Based on the prototype of the classroom, it was found that the AI-powered textbook can enhance students' practical English learning by providing more diagnostic, adaptive and evidence-based textbook content within a stable curriculum structure.

The study still has defects. The experiment only included one university and an eight-week teaching period; thus, the results may not be generalizable to extended curriculum application in other institutions. The prototype also only covered some general practical English situations and did not study specialised English for engineering, medicine, business and tourism in detail. Strengthen the governance function of teachers, add batch review, local resource modification and explainable recommendation reports.

Future work will extend the prototype to multi-institution trials, longer learning cycles and more specialised practical English areas. Further work should also improve multimodal data processing, privacy-preserving learner modelling, and teacher-facing analytical explanations. An excellent AI-powered practical English textbook should not just be an intelligent addition to the existing materials; rather, it should be a transparent, manageable and pedagogically sound platform for learning.